\documentclass[conference]{IEEEtran}
\IEEEoverridecommandlockouts 

\usepackage{graphicx} 
\usepackage{cite}
\usepackage{amsmath,amssymb,amsfonts}
\usepackage{algorithmic}
\usepackage{graphicx}
\usepackage{textcomp}
\usepackage{xcolor}
\def\BibTeX{{\rm B\kern-.05em{\sc i\kern-.025em b}\kern-.08em
    T\kern-.1667em\lower.7ex\hbox{E}\kern-.125emX}}

\usepackage{soul}

\begin{document}

\title{
Federated Attention Autoencoders with a Stochastic Aggregation Scheme for Anomaly Detection
\thanks{\fontsize{7.5pt}{8.5pt}\selectfont \copyright~2025 IEEE. Personal use of this material is permitted. Permission from IEEE must be obtained for all other uses, in any current or future media, including reprinting/republishing this material for advertising or promotional purposes, creating new collective works, for resale or redistribution to servers or lists, or reuse of any copyrighted component of this work in other works.}%
}

\author{
\IEEEauthorblockN{
Mihailo Ilić\IEEEauthorrefmark{1},
Miloš Savić\IEEEauthorrefmark{1},
Vladimir Kurbalija\IEEEauthorrefmark{1},
Mirjana Ivanović\IEEEauthorrefmark{1},
Giancarlo Fortino\IEEEauthorrefmark{2},
Dušan Jakovetić\IEEEauthorrefmark{1}
}

\\

\IEEEauthorblockA{\IEEEauthorrefmark{1}%
Department of Mathematics and Informatics, Faculty of Sciences, University of Novi Sad, Serbia}
\{milic,svc,kurba,mira,dusan.jakovetic\}@dmi.uns.ac.rs
\IEEEauthorblockA{\IEEEauthorrefmark{2}%
Department of Informatics, Modeling, Electronics, and Systems, University of Calabria, Italy}
giancarlo.fortino@unical.it
}

\maketitle
\IEEEpubidadjcol

\begin{abstract}
Outlier detection in decentralized data environments is a challenging task for many machine learning implementations, particularly in settings where data cannot be shared. Recently, there have been advances in federated outlier detection, some of which are based on the use of autoencoder networks. The introduction of attention mechanisms to autoencoders boosts their efficiency. However, the application of attention-based models in federated learning remains underdeveloped due to the absence of proper aggregation functions for these types of networks. In our work, we propose two novel aggregation functions tailored for attention-based autoencoders, which better preserve the learned information stored within the memory modules of these networks. We evaluated our approach on the KDDCUP10 dataset, and we showed that the proposed methods achieve up to 2.9\% and 5.1\% better results for F1 score and AUC ROC respectively when compared to traditional autoencoders.
\end{abstract}


\section{Introduction}

Federated learning (FL)~\cite{kairouz2021advances} has emerged as a decentralized learning paradigm for training machine learning (ML) models across multiple clients (edge nodes). FL is different from standard distributed learning in that data is never centralized and it is collected at the edge. There are numerous benefits to this approach, some of the most important being increased security and privacy, and the reduction of communication overhead needed to centralize the data or to distribute initially centralized data across a cluster of workers. In most cases, the FL process is coordinated by a central server which conducts client selection and model aggregation~\cite{ilic2024towards, qi2024model}. 
After the clients send their local models to the server, an aggregation step follows. The choice of the aggregation function is extremely important~\cite{qi2024model}, as it directly affects the quality of the federated model. 
Furthermore, this choice is influenced by challenges like increasing privacy, reducing communication overhead, computational heterogeneity between devices, energy efficiency, and many others~\cite{kairouz2021advances,thakur2025green}.

Different model aggregation functions, such as FedAvg~\cite{mcmahan2017communication}, FedProx~\cite{li2020federated}, and FedHybrid~\cite{niu2023fedhybrid} can yield substantially different updated global models. The main idea of FL is to train global models which generalize well on the entirety of the distributed dataset. This is one of the main challenges for FL, as no assumptions can be made about the distributions from which the data was drawn at each individual node. Balancing between good generalizability and effectiveness at each edge node is called the model personalization problem. Solutions like client clustering, multi-task learning, and local fine-tuning \cite{li2020review, hanzely2020federated, sattler2020clustered, armacki2022personalized, wang2019federated} have been proposed to solve this problem. Approaches like \cite{deng2021distributionallyrobustfederatedaveraging, mohri2019agnosticfederatedlearning} aim to solve the issue of varying data distributions across clients, providing solutions that are distributionally robust.

Representational learning (RL) is a technique which involves learning how to extract relevant features from inputs to ML models during training and inference. By doing so, generalizable features can be obtained from the original input dataset. RL algorithms learn representations that capture underlying factors, allowing this knowledge to be applicable to multiple learning tasks~\cite{bengio2013representation}. Learning representations can be done in numerous ways, one of them being through the use of autoencoder networks consisting of 2 parts: the encoder and the decoder. The network learns to represent the inputs in a latent space. Representational learning can be done in an unsupervised, semi-supervised, or self-supervised fashion~\cite{saeed2019multi, dosovitskiy2014discriminative, bengio2013representation}. Recently, the use of memory-augmented autoencoders, i.e., those relying on attention mechanisms have also been mentioned in the context of federated representational learning~\cite{gong2019memorizing, anwar2024fedad}.

Anomalies or outliers are data instances in a sample that considerably deviate from other members of the sample or from some concept of normality~\cite{Ruff2021,Nassif2021}. The detection of anomalies is widely used in a broad variety of applications and they can be classified into three categories: point anomalies, contextual anomalies and collective anomalies. A point anomaly is an individual anomalous data instance (e.g., an illegal transaction, an image of a damaged product in manufacturing). A contextual anomaly is a data instance that is anomalous in a specific time, space, or structural context (e.g., time-series, spatial, spatio-temporal, or graph-based anomalies). A collective anomaly is a set of related, consequent, or dependent data instances that are contextually anomalous. A fundamental assumption in anomaly detection is that the data space region where the normal data instances exist can be bounded~\cite{Ruff2021}. 

The main contributions of this paper are:
\begin{itemize}
    \item an analysis of the challenges involved in aggregating memory modules of attention-based autoencoders in federated learning;
    \item the proposal of novel FL aggregation functions for attention-based autoencoders;
    \item an experimental evaluation showing improved performance of said functions compared to using standard aggregation functions like FedAvg~\cite{mcmahan2017communication} or FedProx~\cite{li2020federated}. 
\end{itemize}

The rest of the paper is structured as follows. Section \ref{sec:rw} covers related work on FL, particularly outlier detection and attention-based networks. The approach and novel methods are discussed in Section \ref{sec:methodology}. Experimental results are presented and discussed in Section \ref{sec:results}. Concluding remarks and plans for future work are given in the last section. 

\section{Related Work}
\label{sec:rw}

Depending on the presence of labels that distinguish normal data instances from anomalous ones in a training dataset, anomaly detection models can be unsupervised, semi-supervised or supervised~\cite{Nassif2021}. The authors of~\cite{Ruff2021} emphasize that there are two big classes of anomaly detection approaches: shallow and deep anomaly detection methods. They can be viewed from a unified perspective, depending on the type of anomaly detection model that belongs to one of the following three categories: 

\begin{enumerate}
    \item density estimation and probabilistic models, 
    \item one-class classification models, and
    \item reconstruction models.
\end{enumerate}

Density estimation and probabilistic models predict anomalies through the estimation of normal data probability distributions. Those methods include classical density estimation, energy-based models, and neural generative models. One-class classification is a discriminative approach to anomaly detection with the goal of learning a decision boundary that separates normal from outlier data instances. Reconstruction models are based on learning low-dimensional data representations of normal data instances. Typical approaches are different variants of principal component analysis, prototypical clustering, and autoencoders.

The work in~\cite{Pang2021} indicates the main challenges and major issues in contemporary anomaly detection: low anomaly detection recall rate, anomaly detection in high-dimensional data, data-efficient learning of normality, noise-resilient anomaly detection, detection of complex anomalies, and anomaly explanation. The authors also categorized existing deep learning anomaly detection methods into three groups: 
\begin{enumerate}
    \item generic normality feature learning (e.g., autoencoders, generative adversarial networks), 
    \item measure-dependent normality feature learning (e.g., clustering, one-class classification), and 
    \item end-to-end anomaly score learning (e.g., anomaly ranking models). 
\end{enumerate}
The article also emphasizes self-supervised learning approaches for generic normality feature learning based on the assumption that normal data points are more consistent to self-supervised learning than anomalies.

Unsupervised anomaly detection in federated settings has numerous important applications related to the security and robustness of Internet of Things (IoT) systems, as they generate enormous amounts of data. Consequently, managing these big data systems brings its own challenges in terms of security, ownership, and privacy~\cite{DBLP:journals/comsur/AwayshehAGNV21}, which FL can help alleviate. The authors of~\cite{nguyen2019} have proposed a self-learning anomaly detection method for intrusion detection in IoT systems. This method uses federated learning to efficiently aggregate anomaly detection profiles. A framework using federated learning to detect malware affecting IoT devices was introduced in~\cite{Rey2022}. The authors used both unsupervised and supervised anomaly detection models showing that they achieve comparable performance to their centralized counterparts while preserving privacy (due to the privacy-by-design nature of FL). The work presented in~\cite{Wang2022} investigated federated anomaly detection for finding battery failures in energy storage systems. Besides IoT, federated anomaly detection methods were also applied to sensitive financial data~\cite{herurkar2024}. The authors of~\cite{li2019} addressed the problem of abnormal client detection in horizontal federated learning systems (such clients could be malicious or malfunctioning). Their solution is based on low-dimensional surrogates of model weight vectors that are used to perform anomaly detection at the server side.
Adequate client selection may also be carried out through zero-trust approaches like in~\cite{DBLP:journals/jsac/TahirMAAGA25}. In this approach, clients are ranked based on multiple criteria, following which the top $N$ are selected to participate in learning in each training round. This kind of technique could also be coupled with anomaly detection mechanisms.

Federated representation learning (FRL) joins federated and representational learning~\cite{Berlo2020}. The motivation behind FRL lies in the fact that data distributions between learning clients can differ substantially.
Non-IID data can cause substantial issues in federated learning scenarios~\cite{10361408,10840118}. Thus, learning a representation of a distributed dataset can increase the performance of collaboratively trained models, including also models for anomaly detection. Having in mind that FL models are typically deep neural networks, since the aggregation of neural network weights is relatively straightforward, autoencoders are the most natural choice for FRL models~\cite{Polato2021,Zhang2023,NovoaParadela2023}.

Besides providing low-dimensional data representations, autoencoders can also be used for unsupervised anomaly detection following the idea that outliers will have larger autoencoder reconstruction errors than inliers or normal data points~\cite{Chen2018,Savic2021,SAVIC2022}. The work in~\cite{anwar2024fedad} indicated key research directions and empirically investigated various unsupervised anomaly detection approaches in federated settings. In the conducted experiments, the authors evaluated five different models on four different datasets. The examined models include deep autoencoders~\cite{Chen2018}, deep structured energy-based models~\cite{Zhai2016}, deep one-class classification~\cite{pmlr-v80-ruff18a}, neural transformation learning with deep networks~\cite{QiuPKMR21} and memory-augmented deep autoencoders~\cite{gong2019memorizing}. These experiments established a benchmark for validating the performance of anomaly detection models in federated settings. As one of the key future research directions, the authors identified the need for more advanced aggregation functions when training federated unsupervised anomaly detection models.

\section{Methodology}
\label{sec:methodology}

In this paper, we consider outlier detection via autoencoder networks. Autoencoders are general-purpose models, with anomaly detection being only one of their numerous applications. The techniques presented in this paper were validated on this learning task, but can be applied to any other.

\begin{figure}
    \centering
    \includegraphics[width=\linewidth]{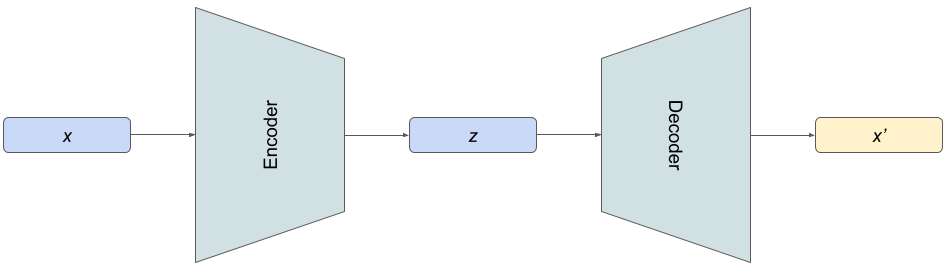}
    \caption{Diagram of an autoencoder network. The encoder part transforms the input vector $x$ into a latent representation $z$. Following this, the decoder produces $x'$ from $z$ in an attempt to reconstruct the input $x$.}
    \label{fig:autoencoder_diagram}
\end{figure}

These networks consist of two components, an encoder and a decoder. The premise behind this approach is to take an input vector $x$, and transform it using the encoder to a different, smaller representation $z$. This representation is also referred to as a \textit{code} or \textit{latent representation}. The decoder uses $z$ as input and outputs $x'$, attempting to reconstruct the original input $x$. A diagram depicting the high-level structure of an autoencoder network can be seen in Figure~\ref{fig:autoencoder_diagram}. The squared $l2$ norm difference between $x$ and $x'$ is called the reconstruction loss, as seen in (\ref{eq:reconstruction_loss}). This type of network yields low reconstruction loss values for inputs similar to those it was trained on. If the loss is too high, the input vector $x$ can be considered an outlier. 

\begin{equation}
    \label{eq:reconstruction_loss}
    e = ||x - x'||_2^2
\end{equation}

\begin{figure}[htb!]
    \centering
    \includegraphics[width=\linewidth]{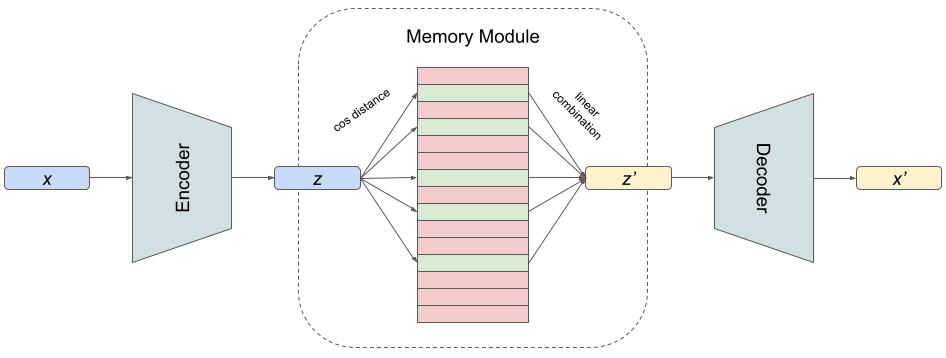}
    \caption{Diagram of the MemAE model. The encoder outputs a latent representation $z$ of input $x$. This vector, along with the memory module, produce a new representation $z'$, which is fed into the decoder to produce the output.}
    \label{fig:memae_diagram}
\end{figure}

The research presented in this paper focuses on a modified variant of autoencoders, called memory-augmented autoencoders, or MemAE~\cite{gong2019memorizing}. These autoencoders utilize an attention mechanism, implemented as a memory module between the encoder and decoder parts of the network, a diagram of which can be seen in Figure~\ref{fig:memae_diagram}. The memory module is a matrix of size $N \times C$, designed to store representative examples of the latent representations of the dataset on which it was trained on. $N$ represents the number of examples stored in the memory module, while $C$ is the size of each encoded example. Once encoded into $z$, a new data point is compared with all the examples stored in the memory module. 
A new code $z'$ is constructed via a linear combination of the representatives most similar to the new data point $z$ after which $z'$ is fed into the decoder. The idea behind this approach is to additionally penalize examples not seen during training, having them produce a larger reconstruction loss. A detailed description of this approach can be found in \cite{gong2019memorizing}, along with experimental results in centralized learning scenarios. 

The authors of \cite{anwar2024fedad} conducted research on the use of different outlier detection models, one of which was MemAE. In their experiments, the MemAE model performed quite well over four experimental datasets with varying hyperparameters, compared to regular autoencoders or even some other advanced methods. Nevertheless, it is pointed out in \cite{anwar2024fedad} that structurally specific elements, like the memory module of MemAE, are not propagated 
during the aggregation process of FL. The authors argue that this is a limitation to the effective use of this approach, as specific heuristics in the memory module are not aggregated.

In this work, we aim to introduce novel aggregation methods for attention-based autoencoder networks. 

\subsection{Memory Matrix Aggregations}
\label{sec:metodology-mem-matrix}

Neural networks are ML models commonly used in FL settings. Because of this, various aggregation functions, like FedAvg~\cite{mcmahan2017communication} and FedProx~\cite{li2020federated}, have been developed which handle neural network weights received from edge nodes, and aggregate them into the global model weights based on various importance metrics. 

The memory module of MemAE is a non-typical element of a neural network. This attention mechanism is a matrix of dimensions $N \times C$ where $N$ is the number of example codes the memory module should store, while $C$ represents the dimension of the encoded vectors. Intuitively, the rows in the matrix store representative results of the training dataset used to tune the model. This matrix is constructed in such a way as to minimize the entropy between the particular vectors stored in its rows.
The idea is to obtain a broad set of representative vectors, which can cover the distribution of the training dataset well. 

\begin{figure}
    \centering
    \includegraphics[width=\linewidth]{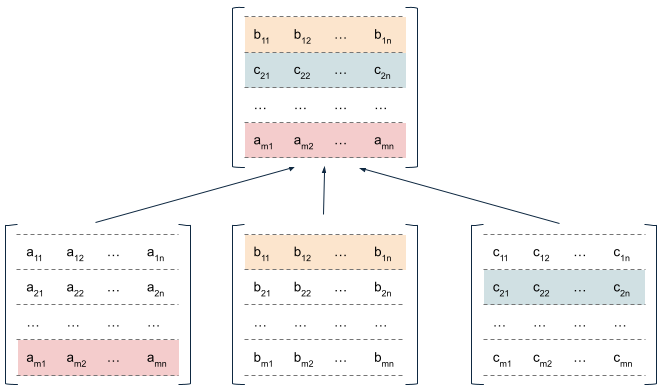}
    \caption{New global memory matrix aggregation via random selection from matrices submitted by edge nodes during one training round. By preserving the rows containing representative codes, no information stored within them is lost.}
    \label{fig:matrix}
\end{figure}

In the experiments conducted in this paper, we propose and validate two new aggregation strategies which take into account both the network weights and the memory module:

\begin{enumerate}
    \item \textbf{Naive aggregation --} FedAvg and similar methods apply a variant of weighted averaging over the weights of the networks obtained from edge nodes. In our first approach, we propagate both the weights and the memory module to the FL coordinator, after which the set of memory modules is treated the same as the set of network weights and is aggregated using a weighted average modified by additional proximal terms via FedProx~\cite{li2020federated}.
    \item \textbf{Random row permutations --} Information is stored in the rows of the matrix of the memory module. In this approach, an aggregated memory module is formed by randomly selecting rows from all the matrices received from the edge nodes during one round of training. The rows that form the newly aggregated memory matrix are taken uniformly at random from the memory modules belonging to the clients, which is illustrated in Figure~\ref{fig:matrix}.
\end{enumerate}

The naive aggregation method simply extends regular aggregation methods like FedProx, having them treat the memory module like any other weight in the neural network. 
Previously in related work~\cite{anwar2024fedad}, the memory module was not propagated during federated training. In our proposed naive aggregation method, the memory module is also shared with the FL coordinator, and is aggregated using the same approach as all other network weights are, i.e. using some sort of weighted average.
However, this approach does not take into account that information in these matrices is stored in their rows, one representative of the dataset in each row. Thus, the order of the rows (vectors) in each matrix received from the edge nodes would influence the resulting aggregated matrix. Just by swapping two rows in a single edge matrix, the resulting global matrix would be different. The second approach takes into account the nature of the data being stored in the memory module and aims to preserve the information learned from the edge nodes. The FL server simply selects a random subset of rows from each edge node, as seen in Figure \ref{fig:matrix}, and stacks them to form a new matrix of the same dimensions. This new matrix contains specific encoded vectors learned from the distributed dataset, and is then used as a memory module in the new global model, which is distributed to the edge nodes. The rest of the weights are aggregated using a standard approach, which in our case is FedProx.

\subsection{The Dataset}

The experiments in this paper were conducted on the KDDCUP10~\cite{kdd_cup_1999_data_130} network intrusion dataset, which is a 10\% subset of the original KDDCUP dataset. It contains 494021 data samples, with 42 different continual and categorical attributes, including the class label. 
The classes cover normal network traffic and 4 categories of network intrusion attacks: denial of service, unauthorized access from a remote machine, unauthorized access to root privileges, and probing or surveillance attacks.
Following data preprocessing, the final number of attributes was 115, excluding the class. 

All of the network intrusion attacks are merged into a single class, opposed to regular network usage. This gives an outlier ratio of 19.69\%. 

\begin{figure}
    \centering
    \includegraphics[width=\linewidth]{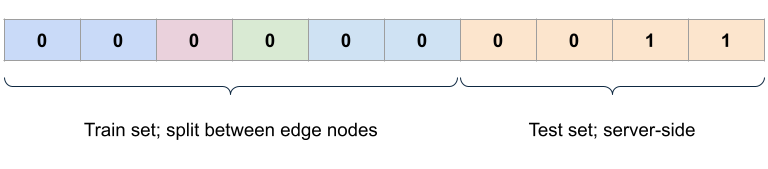}
    \caption{The test set contains a 50-50 split of inlier (marked \textit{"0"}) vs outlier (marked \textit{"1"}) data points. The leftover datapoints are all inliers, and are distributed randomly between edge nodes. Local datasets are used only for training purposes.}
    \label{fig:data_split}
\end{figure}

The dataset was split into training and test sets. The test set contains all outlier data points. A random selection of inliers is added to the test set, forming a 50-50 split of inliers and outliers. The remaining inlier data is then split into $E$ different subsets for training, where $E$ represents the number of edge nodes in the system. This process can be seen in Figure~\ref{fig:data_split}. In our experiments, each of the $E$ edge nodes contained the same amount of data. 

\subsection{Experimental Setup}

Two types of models were used in these experiments: regular autoencoders and memory-augmented autoencoders (MemAE). 

The experiment covered multiple scenarios:

\begin{enumerate}
    \item Independent local training -- Each of the edge nodes trains its own model, using data available locally. No parameter sharing (FL) is involved in this process. In this scenario, regular autoencoders were trained;
    \item Autoencoder-based FL -- Edge nodes train collaboratively through FL. The model in question is a regular autoencoder;
    \item MemAE FL with naive aggregation;
    \item MemAE FL with random row permutations.
\end{enumerate}

The first two experiments are used as baselines to which the newly proposed aggregation strategies were compared. Independent local training serves as the first baseline to verify the motivation for using FL in these scenarios, while the second experiment represents the baseline FL strategy. 

In each of the FL experiments, FedProx was used as a base aggregation strategy. 
This means that the neural network weights for all scenarios, 2 through 4, were aggregated using this aggregation function. In scenario 3, FedProx was also used to aggregate the memory module (naive aggregation), while in scenario 4, the random row permutation strategy was used to aggregate the memory module.

\begin{table}[htbp]
    \caption{Hyperparameters of the FL Experiments}
    \begin{center}
        \begin{tabular}{|c|c|c|}
            \hline
            \textbf{Parameter} & \textbf{Value for $E = 3$} & \textbf{Value for $E = 50$} \\
            \hline
            Fraction Fit & 100\% & 10\% \\
            \hline
            Batch Size & 64 & 1024 \\
            \hline
            \# Global Training Rounds & \multicolumn{2}{|c|}{10} \\
            \hline
            FedProx Proximal $\mu$ & \multicolumn{2}{|c|}{0.01} \\
            \hline
            \# Local Epochs & \multicolumn{2}{|c|}{5}  \\
            \hline
            Optimizer & \multicolumn{2}{|c|}{Adam}  \\
            \hline
            Learning Rate & \multicolumn{2}{|c|}{0.01}  \\
            \hline
            Memory Dimension ($N$) & \multicolumn{2}{|c|}{50} \\
            \hline
            Encoding Size ($C$) & \multicolumn{2}{|c|}{10} \\
            \hline
        \end{tabular}
        \label{tab:exp_params}
    \end{center}
\end{table}

Two sets of experiments were conducted: in the first, the number of edge nodes $E$ was set to $3$, while in the second, $E = 50$. 
All edge nodes participated in the training process in the first set of experiments. During federated training in the second set of experiments, a fraction of edge nodes chosen uniformly at random participated in each global training round, as seen in Table~\ref{tab:exp_params}. 

At the end of each global training round, after model aggregation, the newly created global model was validated against the test set on the server side. In this approach, we aim to train a robust global model capable of generalizing well across the distributed environment. In independent local training, each edge node validates its model against the shared test set.

Validation was done by applying min-max normalization on all of the reconstruction losses for data points in the test set. 
Following this, the threshold value that maximizes the F1 score based on the precision-recall curve was selected, in order to compare the best possible models in each case.

\section{Results and Discussion}
\label{sec:results}

In each experiment, all participating edge nodes contributed exactly 50 epochs of training. The federated learning experiments lasted for 10 global training rounds, and regardless of client selection, each edge node trained for 5 epochs in one round. In independent local experiments, all edge nodes trained their own models independently for 50 epochs. After each epoch in local training and each global training step during federated training, the trained model was validated on the centralized test set with a 50-50 inlier-to-outlier ratio. For each experiment, the model with the highest F1 score was taken as the best during that training round, and these are the results shown in Tables~\ref{tab:max_vals_e_3} and~\ref{tab:max_vals_e_50}. This was done in order to compare the best possible models in each of the scenarios.  

Table~\ref{tab:max_vals_e_3} depicts the experiments carried out on 3 edge nodes. The row permutation function produced the best aggregated model in terms of both F1 and AUC ROC scores. This approach yielded an F1 score of $98.2\%$, which is significantly higher than the regular autoencoder approach, which is $95.3\%$. The naive aggregation method was also scored higher than the autoencoder, at $97.6\%$. The local independent models in this experiment significantly underperformed, with the best one achieving an F1 score of only 87.8\%. 

\begin{table}[htbp]
    \caption{Maximum Recorded Metric Values when E = 3}
    \begin{center}
        \begin{tabular}{|c|c|c|c|c|}
            \hline
            \textbf{Experiment} & \textbf{Precision} & \textbf{Recall} & \textbf{F1 Score} & \textbf{AUC ROC} \\
            \hline
            Local Independent & \textbf{99} & 78.9 & 87.8 & 84.2 \\
            \hline
            Regular Autoencoder & 98.8 & 92.1 & 95.3 & 93.8 \\
            \hline
            Naive Aggregation & 97.1 & \textbf{98.2} & 97.6 & 97.8 \\
            \hline
            Row Permutation & 98.5 & 97.8 & \textbf{98.2} & \textbf{98.9} \\
            \hline
        \end{tabular}
        \label{tab:max_vals_e_3}
    \end{center}
\end{table}

Experiments with 50 edge nodes show a significant increase in the performance of local models, which can be seen in Table~\ref{tab:max_vals_e_50}. The best local model (out of 50) is tied with the federated model aggregated via the row permutation approach, in terms of F1 measure, at $98.9\%$. However, the federated model is slightly better in terms of AUC ROC, by a margin of $0.1\%$. The autoencoder performed worse than the row permutation approach with the MemAE model, with an F1 score of $96.2\%$. The naive aggregation method drastically underperformed with the highest recorded F1 score of only $85\%$. This can be attributed to the approach in aggregating the memory matrix, where it is treated like other weights in the network and aggregated in the same way as them. The nature of the data stored within the memory module is not taken into account by this approach, as discussed in Section~\ref{sec:metodology-mem-matrix}.

\begin{table}[htbp]
    \caption{Maximum Recorded Metric Values when E = 50}
    \begin{center}
        \begin{tabular}{|c|c|c|c|c|}
            \hline
            \textbf{Experiment} & \textbf{Precision} & \textbf{Recall} & \textbf{F1 Score} & \textbf{AUC ROC} \\
            \hline
            Local Independent & 98.3 & \textbf{99.6} & \textbf{98.9} & 99.3 \\
            \hline
            Regular Autoencoder & \textbf{98.6} & 93.8 & 96.2 & 97 \\
            \hline
            Naive Aggregation & 76.7 & 95.4 & 85 & 89.7\\
            \hline
            Row Permutation & 98.5 & 99.3 & \textbf{98.9} & \textbf{99.4}\\
            \hline
        \end{tabular}
        \label{tab:max_vals_e_50}
    \end{center}
\end{table}

As seen in Table~\ref{tab:max_vals_e_50}, the best independent model achieved the same results as the row permutation based federated network. Since 50 different local models have been trained independently, it is worth exploring the performance of models from individual edge nodes. Table~\ref{tab:statistics_e_50} shows the aggregated statistics for all 50 edge nodes. The best F1 score is $98.9\%$, while the worst is $79.7\%$. In the 50th percentile, the F1 score is at $97.4\%$, which is still better than the regular autoencoder network trained in a federated fashion. However, the 75th percentile shows that most of the edge nodes trained models which underperformed compared to the row permutation approach. Some of the clients managed to train good models, however, a lot of them would have benefited from federated training, particularly from the row permutation approach. The mean F1 score was $95.3\%$, which is significantly lower than the row permutation approach, and is also outperformed by the basic federated autoencoder.

\begin{table}[htbp]
    \caption{Local Training Statistics E = 50}
    \begin{center}
        \begin{tabular}{|c|c|c|}
            \hline
            \textbf{Statistic} & \textbf{F1 Score} & \textbf{AUC ROC} \\
            \hline
            Mean & \textbf{95.3} & \textbf{95.1} \\
            \hline
            Min & 79.7 & 78.2 \\
            \hline
            25\% & 94.2 & 93.7\\
            \hline
            50\% & 97.4 & 97.9\\
            \hline
            75\% & 98.1 & 99.1\\
            \hline
            Max & 98.9 & 99.5\\ 
            \hline
        \end{tabular}
        \label{tab:statistics_e_50}
    \end{center}
\end{table}

\begin{figure}
    \centering
    \includegraphics[width=\linewidth]{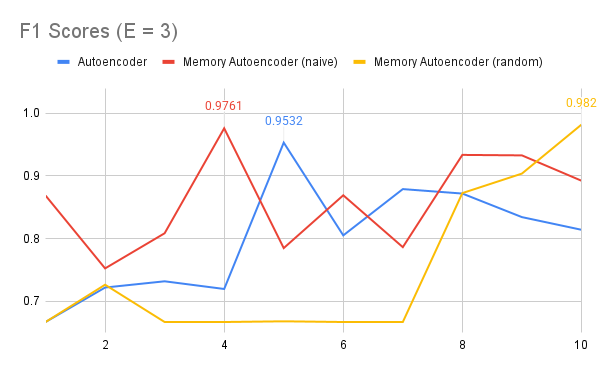}
    \caption{F1 scores through 10 global rounds of federated training on 3 edge nodes.}
    \label{fig:f1_3}
\end{figure}

\begin{figure}
    \centering
    \includegraphics[width=\linewidth]{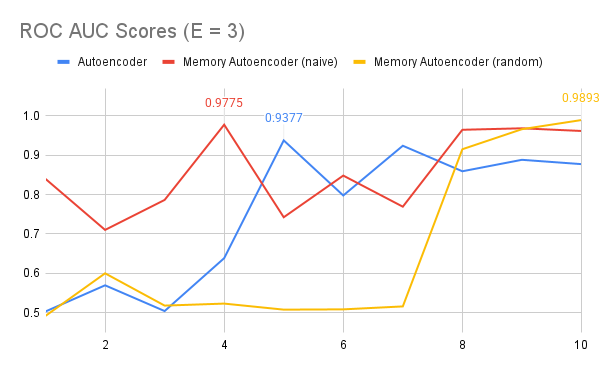}
    \caption{AUC ROC scores through 10 global rounds of federated training on 3 edge nodes.}
    \label{fig:auc_roc_3}
\end{figure}

Figures~\ref{fig:f1_3} and~\ref{fig:auc_roc_3} show the changes in F1 and AUC ROC scores during the 10 rounds of federated training on 3 edge nodes. The regular autoencoder and naive aggregation MemAE achieved their best results in the first half of training. The random permutation approach was more stable in that regard, having consistent values in the early stages of training and, by the end, yielding the best performing models. 

\begin{figure}
    \centering
    \includegraphics[width=\linewidth]{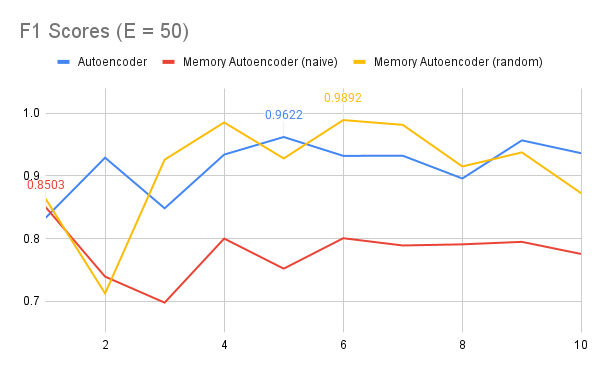}
    \caption{F1 scores through 10 global rounds of federated training on 50 edge nodes.}
    \label{fig:f1_50}
\end{figure}

\begin{figure}
    \centering
    \includegraphics[width=\linewidth]{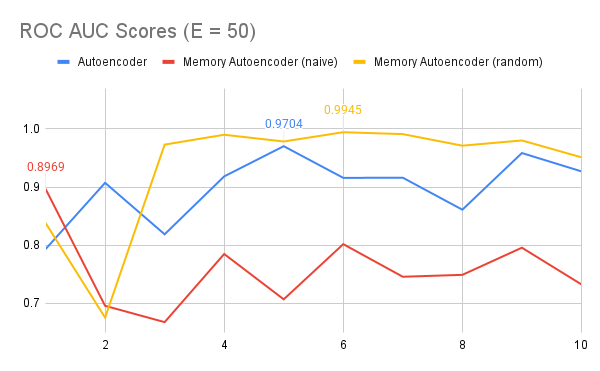}
    \caption{AUC ROC scores through 10 global rounds of federated training on 50 edge nodes.}
    \label{fig:auc_roc_50}
\end{figure}

The training process on 50 edge nodes produced the same end result, with a different training dynamic. Figures~\ref{fig:f1_50} and~\ref{fig:auc_roc_50} show that random aggregation yielded the best model somewhere in the middle of training. The naive approach dropped off immediately, further emphasizing the importance of handling the intricacies of the memory module carefully. The regular autoencoder achieved solid results, however, it still underperformed compared to the row permutation approach.  
 
The data shows that it is possible to obtain high quality models by using well-defined aggregation functions that take into account the specifics of the memory-augmented autoencoder. This approach outperformed the standard approach of training regular autoencoders. The naive aggregation function demonstrates the importance of taking into account the heuristics of the attention-based model and the importance of handling the latent representations stored in the memory module with care. 
The newly proposed aggregation function based on random row permutations showed considerable improvement over other methods. As a stochastic approach, it uniformly samples encoded representations from memory modules provided by clients. The increased performance comes from the preservation of the learned representations that this approach offers. Extending this approach with a guided process of selecting representative examples from the memory modules could further improve its robustness and results.  
One such way is to make use of submodular optimization~\cite{lin2009select, lin2010application, iyer2013submodular}, and selecting different representatives based on the amount of new information they carry. This would optimize the process of aggregating the memory matrix. 

\section{Conclusion}

Federated learning is currently generating a lot of interest from both academia and industry. Some work has already been done on the use of different techniques for outlier detection in federated settings. However, the topics of federated representational learning and outlier detection are still fairly unexplored, with room for different research directions.

The results presented in this paper support the notion that row-preserving aggregation functions for attention-based autoencoders yield better results when training in federated environments. Contextual information is stored in the latent representations kept in the memory modules of individual edge nodes, and it is crucial to keep them intact in order to preserve the knowledge gained during training. Information is destroyed when simply propagating the memory module and aggregating it in the same way as neural network weights. 

The findings in this paper show promising initial results validated on the KDDCUP dataset. Broader validation is needed with different baseline aggregation algorithms, along with different datasets and ML tasks. Future work also includes the validation of the aggregation techniques proposed in this paper on additional data modalities (i.e., time-series). Furthermore, a more extensive and formal analysis of the random permutation approach is needed, as it is a stochastic process.

We also plan to develop new aggregation techniques based on submodular optimization~\cite{lin2009select, lin2010application, iyer2013submodular}. This technique helps when selecting examples from a dataset that minimize redundancy with each other, which could lead to a more optimal selection of representative encodings of the federated dataset. 

\section*{Acknowledgment}

This paper has been supported by the European Union’s Horizon Europe research and innovation actions under grant agreement No 101135775 (PANDORA).

The authors from the University of Novi Sad gratefully acknowledge the financial support of the Ministry of Science, Technological Development and Innovation of the Republic of Serbia (Grants No. 451-03-137/2025-03/ 200125 \& 451-03-136/2025-03/ 200125).

\bibliographystyle{abbrv}
\bibliography{ref.bib}

\end{document}